\documentclass[runningheads]{llncs}
\usepackage{graphicx}

\usepackage{tikz}
\usepackage{comment}
\usepackage{amsmath,amssymb} 
\usepackage{color}
\usepackage{bbm}

\usepackage[accsupp]{axessibility}  

\usepackage{color}

\usepackage{caption}
\usepackage{subcaption}
\usepackage{orcidlink}

\begin{document}
\pagestyle{headings}
\mainmatter
\def\ECCVSubNumber{}  

\title{Number-Adaptive Prototype Learning for \\ 3D Point Cloud Semantic Segmentation}

\titlerunning{NAPL for 3D Point Cloud Semantic Segmentation}
%
\author{Yangheng Zhao\inst{1}\orcidlink{0000-0002-1467-2989} \and
Jun Wang\inst{2}\orcidlink{0000-0003-1543-5456} \and
Xiaolong Li\inst{3}\orcidlink{0000-0002-0662-4311} \and
Yue Hu\inst{1}\orcidlink{0000-0002-1125-8897} \and
Ce Zhang\inst{3}\orcidlink{0000-0002-4344-5259} \and 
Yanfeng Wang \inst{1,4} \and
Siheng Chen \inst{1,4}\orcidlink{0000-0001-6199-529X} 
\\[.21cm]
$^{1}$ Shanghai Jiao Tong University \quad
$^{2}$ University of Maryland \\
$^{3}$ Virginia Tech \quad
$^{4}$ Shanghai AI laboratory \\
\tt\small \email{\{zhaoyangheng-sjtu,18671129361,wangyanfeng,sihengc\}@sjtu.edu.cn} \email{junwang@umiacs.umd.edu} \email{\{lxiaol9,zce\}@vt.edu}
}
\authorrunning{Y. Zhao et al.}
%
\institute{}
\maketitle

\begin{abstract}
3D point cloud semantic segmentation is one of the fundamental tasks for 3D scene understanding and 
has been widely used in the metaverse applications. 
Many recent 3D semantic segmentation methods learn a single prototype (classifier weights) for each semantic class, and classify 3D points according to their nearest prototype.  However, learning only one prototype for each class limits the model's ability to describe the high variance patterns within a class.  Instead of learning a single prototype for each class, in this paper, we propose to use an adaptive number of prototypes to dynamically describe the different point patterns within a semantic class. 
With the powerful capability of vision transformer, we design a \textit{Number-Adaptive Prototype Learning (NAPL)} model for point cloud semantic segmentation. To train our NAPL model, we propose a simple yet effective \textit{prototype dropout} training strategy, which enables our model to adaptively produce prototypes for each class. The experimental results on SemanticKITTI dataset demonstrate that our method achieves 2.3\% mIoU improvement over the baseline model based on the point-wise classification paradigm. 
\end{abstract}

\section{Introduction}
\begin{figure}[t]
    \centering
    \includegraphics[width=10cm]{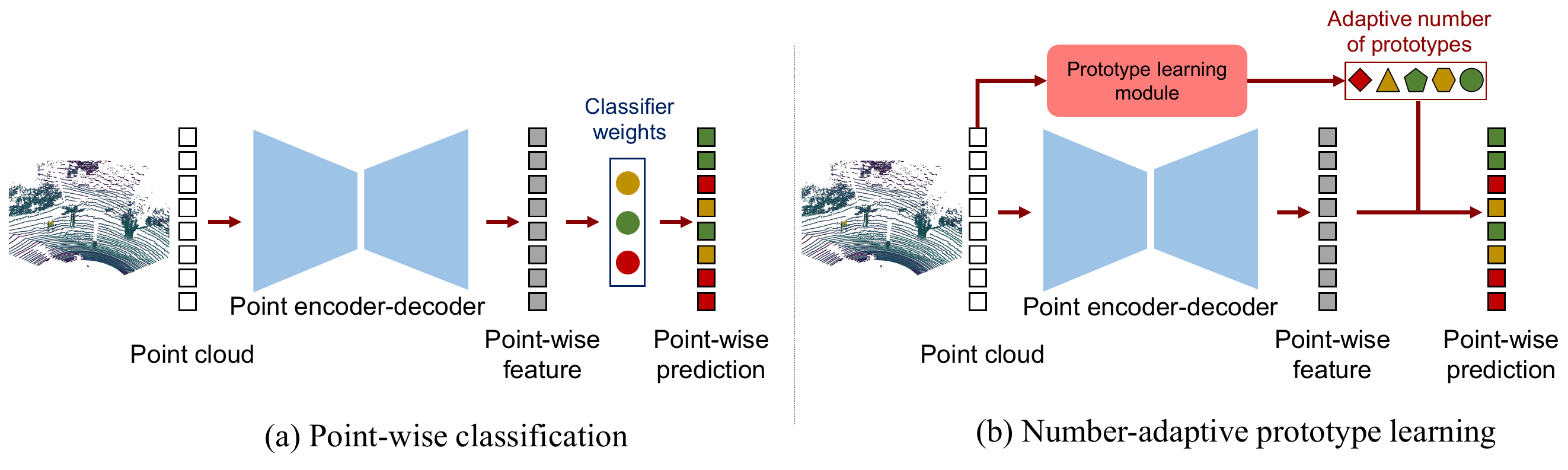}
    \caption{Difference between point-wise classification (PWC) and number-adaptive prototype learning (NAPL).
    The PWC paradigm~\cite{hu2020randla,milioto2019rangenet++,zhu2021cylindrical} learns a single prototype (classifier weights) for each class, while  our proposed NAPL uses a prototype learning module to adaptively produce multiple prototypes for each class.
    }
    \label{fig:fig1}
\end{figure}

3D scene understanding is critical for numerous applications, including metaverse, digital twins and robotics~\cite{chen20203d}. As one of the most important tasks for 3D scene understanding, point cloud semantic segmentation provides point-level understanding of the surrounding 3D environment and gets increasing attention.

A popular paradigm for 3D point cloud semantic segmentation follows the point-wise classification, where an encoder-decoder network extracts point-wise features and feeds them into a classifier predicting label, as shown in Fig.~\ref{fig:fig1} (a). Following the spirit of prototype learning in image semantic segmentation~\cite{zhou2022rethinking}, the point-wise classification model can be viewed as learning one prototype (classifier weights) for each semantic category, and assigning points with the label of the nearest prototype. 
However, the common single-prototype-per-class design in point-wise classification models limits the model's capacity in the semantic categories with high intra-class variance. 
More critically, the 3D point cloud data we are interested in is sparse and non-uniform. The issues of distance variation and the occlusion in 3D point cloud can make the geometric characteristics of objects of the same category very different, and this challenge is even more significant in large-scale 3D data. Experiments show that one prototype per class is usually insufficient to describe those patterns with high variations; see Fig.~\ref{fig:example}. 

To better handle the data variance, an intuitive idea is to use more than one prototype for each category. However, we have no prior knowledge about how many prototypes each category needs, and too many prototypes per category may increase the computational costs while also lead to potential overfitting issues. The question is -- can we find a smarter way to identify the necessary prototypes and effectively increase existing models' capacity? In this work, we propose to use an adaptive way to set the number of prototypes per semantic category, as shown in Fig.~\ref{fig:fig1} (b). We call this paradigm as Number-Adaptive Prototype Learning (NAPL).
To instantiate the proposed NAPL model, inspired by the recent work~\cite{cheng2021per,strudel2021segmenter}
, we use a transformer decoder to learn adaptive number of prototypes for each category. Unlike previous work~\cite{cheng2021per,strudel2021segmenter}, which is limited by learning one prototype for each semantic category, we design a novel \textit{prototype dropout} training strategy, to enable the model adaptively produce prototypes for each class. The experimental results on SemanticKITTI~\cite{behley2019semantickitti} dataset show that 
by plugging our design to a common encoder-decoder network,
our method achieves a 2.3\% mIoU gain than the baseline point-wise classification model.

\section{Related work}

\subsection{3D point cloud semantic segmentation}

3D point cloud semantic segmentation has been widely used in metaverse, digital twins, robotics and autonomous driving~\cite{chen2019suma++,wu2020motionnet}.
Based on different representations, existing 3D semantic segmentation methods can be divided into three categories: projection-based, point-based and voxel-based. The projection-based methods, SqueezeSegV3~\cite{xu2020squeezesegv3} and RangeNet++~\cite{milioto2019rangenet++}, project the 3D point cloud into the 2D plane, and do feature learning and segmentation on the projected 2D image. 
Alternatively, the point-based methods, PointNet ~\cite{qi2017pointnet},  RandLA~\cite{hu2020randla}, KPConv~\cite{thomas2019kpconv}, and PointMotionNet~\cite{wang2022pointmotionnet} learn point-wise features from the raw point cloud with the specifically designed multi-layer perceptron (MLP) and convolution kernels.
The voxel-based methods~\cite{choy20194d,zhu2021cylindrical}, including MinkUNet and Cylinder3D, discretize the space into regular grids, and leverage 3D CNN networks to extract features. In this work, we design our NAPL based on the MinkUNet~\cite{choy20194d}. 

\subsection{Mask classification for image segmentation}
In the 2D image segmentation task, inspired by the pioneering work DETR~\cite{carion2020end}, there is a trend to leverage the mask classification paradigm for semantic segmentation. 
Among those, Segmenter~\cite{strudel2021segmenter} proposes to use a set of learnable queries
to predict class masks for semantic segmentation. 
MaskFormer~\cite{cheng2021per} proposes to unbind the queries from categories, and uses the learnable queries to predict mask embeddings for mask prediction and mask classification. Our model adapts the mask prediction paradigm of MaskFormer from 2D image segmentation to 3D point cloud semantic segmentation. Jointly with the proposed prototype dropout training strategy, our model produces an adaptive number of prototypes for one class, which naturally meets the necessity of 3D semantic segmentation.

\section{Methodology}\label{sec:method}
\begin{figure}[t]
    \centering
    \includegraphics[width=10.0cm]{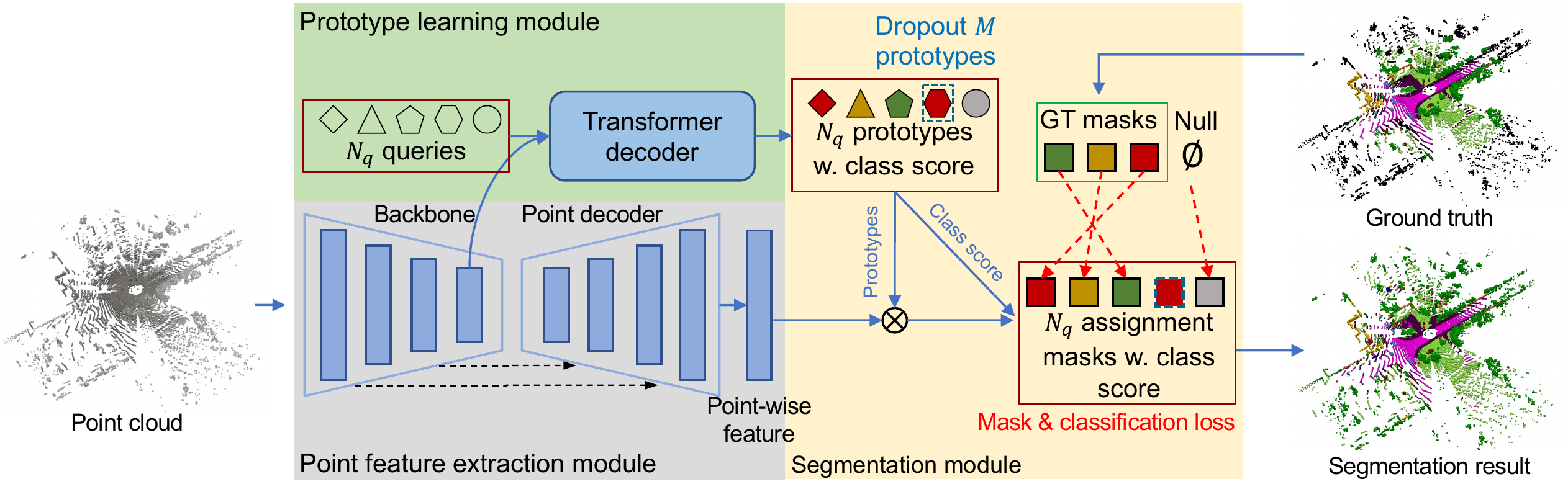}
    \caption{The proposed NAPL consists of three parts: point feature extraction module, prototype learning module, and segmentation module. During training, we randomly drop $M$ prototypes to enable NAPL to adaptively produce prototypes.}  
    \label{fig:NAPL}
\end{figure}

In this section, we first revisit the current 3D semantic segmentation paradigm. We then introduce the proposed number-adaptive prototype learning for 3D point cloud semantic segmentation and a novel strategy to train our model. 

\subsection{Overview of 3D semantic segmentation paradigm}
Given a frame of 3D point cloud $X \in \mathbb{R}^{N\times3}$ with $N$ points, the goal of 3D point semantic segmentation is to predict a semantic class label $c \in \{1, 2, \dots, C \}$ for each point.  For the point-wise classification paradigm, current models~\cite{hu2020randla,milioto2019rangenet++} comprise of two main parts: i) an encoder-decoder network $\phi(\cdot)$ for point-wise feature extraction, and ii) a classifier $\psi$ to project the point features into the semantic label space. For each point $x_i \in X $, 
its feature $f_{i} = \phi(x_i) \in \mathbb{R}^D$ is fed into $\psi$ for $C$-way classification:
$
    p(c|{i})={\exp \left(w_{c}^{\top} f_{i}\right)} / {\sum_{c^{\prime}=1}^{C} \exp \left(w_{c^{\prime}}^{\top} f_{i}\right)}, 
$
where $p(c|{i}) \in \left[0,1 \right]$ is the probability that the $i$th point belongs to the $c$th class, and $\psi$ is parameterized by ${W} = \left[{w}_1,\dots,{w}_C \right] \in \mathbb{R}^{C\times D}$ with ${w}_c \in \mathbb{R}^D$ a learnable vector for the $c$th class. From a prototype view, the label assignment of point $x_i$ is 
$
    \widehat{c}_i = \underset{c}{\arg \max}\{{w}_{c}^{\top} f_{i}\} _{c=1}^{C},
$
where $w_c$ can be viewed as a prototype of class $c$. 

As mentioned in the introduction, the single prototype per class largely limits the model's ability to describe the high variance pattern within a class.

\subsection{Number-adaptive prototype learning} \label{sec:napl}

Different from the single-prototype-per-class design in the point-wise classification paradigm, we propose to use an adaptive number of prototypes for each class. Our model generally comprises of two parts: i) a point feature extraction module $\phi(\cdot)$ to extract point-wise feature, ii) a prototype learning module $ g(\cdot) $ which takes the point cloud as input, and produces $K$ prototype vectors $P = \left[ p_1, p_2, \dots, p_K \right] \in \mathbb{R}^{K\times D}$ with the corresponding class label $\{l_k \in \{1, 2, \dots, C \}\}_{k=1}^{K}$. It is worth noting that the number of class-$c$ prototypes  $k_c = |\{p_k | c_k = c \}|$ depends on the input point cloud $X$, so the total prototype number $K=\sum_{c=1}^C k_c$ is also input-dependent. Finally, the point $x_i$ is labeled as the class of its nearest prototype, which can be formulated as: 
$
    \widehat{c}_i = c_{k^*}, k^* = \underset{k}{\arg \min } \operatorname{dist}(p_k, f_{i})_{k=1}^{K},
$
where $\operatorname{dist}(\cdot, \cdot)$ is the distance between two vectors. 

\subsection{Model architecture}
We now introduce the implementation details of the proposed NAPL. Fig.~\ref{fig:NAPL} overviews our model, which consists of 3 modules: point feature extraction module, prototype learning module, and segmentation module. 

\textbf{Point feature extraction module.} The point feature extraction module (PFEM) takes the raw 3D point cloud as input and extracts point-wise features. It consists of a backbone network to learn compact point features and a decoder network to predict point-wise features $F = \left[f_1, f_2, \dots, f_N \right]\in \mathbb{R}^{N \times D}$ , where $N$ denotes the number of points, and $D$ denotes the feature dimension. We pre-train a per-point classification backbone to initialize our model's feature extraction.

\textbf{Prototype learning module. }
The core challenge of the number-adaptive prototype learning paradigm is the implementation of the prototype learning module.
Inspired by the recent work~\cite{cheng2021per,strudel2021segmenter} in image segmentation, 
we leverage a transformer decoder as our prototype learning module. It takes $N_q$ learnable query vectors and the intermediate point features as input, and then progressively updates query vectors with point features using attention blocks. Finally, it outputs $N_q$ prototype proposals $P = \left[p_1, p_2, \dots, p_{N_q} \right] \in \mathbb{R}^{N_q \times D}$ and the corresponding prototype class score $S = \left[s_1, s_2, \dots, s_{N_q} \right] \in \left[0, 1\right]^{N_q \times (C+1)}$. The additional class label $C+1$ denotes that the corresponding prototype does not belong to any semantic class, which enables that an adaptive number $K \leq N_q$ of prototypes are kept for segmentation.

\textbf{Segmentation module.} Given the point-wise feature $F$ and prototype proposals $(P, S)$, 
we follow the semantic inference procedure of MaskFormer~\cite{cheng2021per} to predict the semantic label for each point. Specifically, the $i$th point's semantic label $\widehat{c}_i$ is obtained by 
$
    \widehat{c}_i = \underset{c\in \{1,2,\dots,C \} }{\arg\max}\sum_{k=1}^{\small N_q}\operatorname{sigmoid}(f_i^{\top}p_k) \cdot s_k^c,
$
where $f_i \in \mathbb{R}^{D}$ is the $i$th point's feature and $s_k^c$ denotes the $c$ th element of $s_k$.

\subsection{Prototype dropout training strategy}
To enable our model to adaptively produce prototypes for segmentation, we design a simple yet effective training strategy, namely prototype dropout training.

The label assignment in Section~\ref{sec:napl} is not differentiable. To facilitate prototype-learning module training, we formulate the assignment between points and prototypes as a set of soft assignment masks $\{m_k = \operatorname{sigmoid}(F \cdot p_k) \in \left[0,1\right]^{N}\}_{k=1}^{N_q}$, and arrange the model prediction as a set of class-mask pairs $z = \{ (s_k, m_k)\}_{k=1}^{N_q}$. Ground truth semantic labels are arranged as the same class-mask pairs $z^{\text{gt}} = \{ (c_k^{\text{gt}}, m_k^{\text{gt}}) |c_k^{\text{gt}} \in \{ 1, 2, \dots, C\}, m_k^{\text{gt}} \in \{0, 1 \}^{N} \}_{k=1}^{N_{\text{gt}}}$, where the $i$th element of $m_k^{\text{gt}}$ suggests whether the point $x_i$ is assigned to the prototypes of class $c_k^{\text{gt}}$. However, simply padding the ground truth set with "no object" tokens $\varnothing$ will
push $N_{\text{gt}}$ of prototypes to the nearest annotated segments, and the remaining $N_q - N_{\text{gt}}$ prototypes to $\varnothing$, resulting in a degraded solution with one prototype per class. 

To encourage adaptive number of prototypes per class, we randomly drop out the class-mask pairs of $M$ prototypes, with the rest denoted as $z^{\prime}$.
We then pad $z^{\text{gt}}$ to the same size as $z^{\prime}$ and calculate the cross entropy loss and mask loss $\mathcal{L}_{\text{mask}}$ between $z^{\prime}$ and $z^{\text{gt}}$ under a minimal matching $\sigma(\cdot)$ to jointly optimize the prototype class prediction and the point-prototype assignment:
\begin{equation*}
    \mathcal{L}(z^{\prime}, z^{\text{gt}})=\sum_{i=1}^{N_q - M}\left[-\log s^{\prime}_{\sigma(i)}\left(c_{i}^{\text{gt}}\right) + \mathbbm{1}_{\left\{c_{i}^{\text{gt}} \neq \varnothing\right\}} \mathcal{L}_{\text{mask}}\left(m_{i}^{\text{gt}}, m^{\prime}_{\sigma(i)}\right)\right],
\end{equation*}
where for the padded token in $z^{\text{gt}}$, whose class label $c_{i}^{\text{gt}} = \varnothing$, we only calculate the cross entropy loss. 
For simplicity, we use the same $\mathcal{L}_{\text{mask}}$ as DETR~\cite{carion2020end}.

\begin{table*}[t]
\begin{center}
\resizebox{\textwidth}{!}{%
\begin{tabular}{l|c|ccccccccccccccccccc}
\hline
  \textbf{Method} &
  \textbf{mIoU} &
  \rotatebox{90} {car} &
  \rotatebox{90} {bicycle} &
  \rotatebox{90} {motorcycle} &
  \rotatebox{90} {truck} &
  \rotatebox{90} {other-vehicle} &
  \rotatebox{90} {person} &
  \rotatebox{90} {bicyclist} &
  \rotatebox{90} {motorcyclist} &
  \rotatebox{90} {road} &
  \rotatebox{90} {parking} &
  \rotatebox{90} {sidewalk} &
  \rotatebox{90} {other-ground} &
  \rotatebox{90} {building} &
  \rotatebox{90} {fence} &
  \rotatebox{90} {vegetation} &
  \rotatebox{90} {trunk} &
  \rotatebox{90} {terrain} &
  \rotatebox{90} {pole} &
  \rotatebox{90} {traffic-sign} \\ \hline\hline
  \multicolumn{20}{c}{test set} \\ \hline
PointNet~\cite{qi2017pointnet}        & 14.6 & 46.3 & 1.3  & 0.3  & 0.1  & 0.8  & 0.2  & 0.2  & 0.0  & 61.6 & 15.8 & 35.7 & 1.4  & 41.4 & 12.9 & 31.0 & 4.6  & 17.6 & 2.4  & 3.7  \\
RandLANet~\cite{hu2020randla}       & 53.9 & 94.2 & 26.0 & 25.8 & 40.1 & 38.9 & 49.2 & 48.2 & 7.2 & 90.7 & 60.3 & 73.7 & 20.4 & 86.9 & 56.3 & 81.4 & 61.3  & 66.8 & 49.2 & 47.7 \\
KPConv~\cite{thomas2019kpconv}          & 58.8 & 96.0 & 30.2 & 42.5 & 33.4 & 44.3 & 61.5 & \textbf{61.6} & 11.8 & 88.8 & 61.3 & 72.7 & \textbf{31.6} & 90.5 & 64.2 & \textbf{84.8} & 69.2 & \textbf{69.1} & 56.4 & 47.4 \\ 
SqueezeSegv3~\cite{xu2020squeezesegv3}    & 55.9 & 92.5 & 38.7 & 36.5 & 29.6 & 33.0 & 45.6 & 46.2 & 20.1 & 91.7 & 63.4 & 74.8 & 26.4 & 89.0 & 59.4 & 82.0 & 58.7 & 65.4 & 49.6 & 58.9 \\
RangeNet++~\cite{milioto2019rangenet++}      & 52.2 & 91.4 & 25.7 & 34.4 & 25.7 & 23.0 & 38.3 & 38.8 & 4.8  & \textbf{91.8} & 65.0 & 75.2 & 27.8 & 87.4 & 58.6 & 80.5 & 55.1 & 64.6 & 47.9 & 55.9 \\
SalsaNext~\cite{cortinhal2020salsanext}       & 59.5 & 91.9 & \textbf{48.3} & 38.6 & 38.9 & 31.9 & \textbf{60.2} & 59.0 & 19.4 & 91.7 & 63.7 & \textbf{75.8} & 29.1 & 90.2 & 64.2 & 81.8 & 63.6 & 66.5 & 54.3 & \textbf{62.1} \\ 

Ours (NAPL) & \textbf{61.6}	& \textbf{96.6} & 32.3 & \textbf{43.6} & \textbf{47.3} & \textbf{47.5} & 51.1 & 53.9 & \textbf{36.5} & 89.6 & \textbf{67.1} & 73.7 & 31.2 & \textbf{91.9} & \textbf{67.4} & \textbf{84.8} & \textbf{69.8} & 68.8 & \textbf{59.1} & 59.2\\ \hline

\multicolumn{20}{c}{validation set} \\ \hline
PWC &	62.3 &	96.2 &	21.5 &	62.0 &	\textbf{78.6}&	50.8 &	68.5 &	\textbf{87.4}&	0.0&	\textbf{93.9}&	\textbf{51.0}&	\textbf{81.3}&	\textbf{1.2}&	\textbf{90.1}&	59.2&	87.8 &	66.1 &	\textbf{73.9}&	64.3 &	50.0	
 \\

Ours (NAPL) &\textbf{64.6}&	\textbf{97.4}&	\textbf{38.2}&	\textbf{71.5}&	74.3&	\textbf{66.2}&	\textbf{71.1}&	81.6&	0.0&	93.1&	48.4&	80.2&	0.2&	90.0&	\textbf{62.6}&	\textbf{89.0}&	\textbf{68.0}&	77.2&	\textbf{66.8}&	\textbf{52.2} \\

\hline
\end{tabular}%
}
\end{center}
\caption{
Quantitative comparison on SemanticKITTI dataset~\cite{behley2019semantickitti}. The proposed NAPL 
outperforms the recent 3D semantic segmentation methods.}
\label{tab:res-kiti}

\end{table*}

\section{Experiments}

\subsection{Implementation details}

\textbf{Dataset.} SemanticKITTI~\cite{behley2019semantickitti} is a widely used benchmark for 3D semantic segmentation. 
We follow ~\cite{hu2020randla} to use the standard training and validation set splits.

\textbf{Model architecture. }Without loss of generality, we use a MinkUNet~\cite{choy20194d} without classifier as our PFEM, which is a fully convolutional voxel-based model with four stages. The input voxel size is 0.05m. We use the fourth stage feature of PFEM and $N_q = 50$ queries as the input of the prototype learning module. 

\textbf{Training details.} 
We use AdamW optimizer and poly learning rate schedule with an initial learning rate of $10^{-3}$ for transformer and point decoder, and $10^{-4}$ for pre-trained backbone. We set the number of dropout prototypes $M = 10$. 
Our model is trained with batch size of 16 on 4 RTX 3090 GPUs for 20 epochs.

\begin{figure}[t]
    \centering
    \includegraphics[width=11cm]{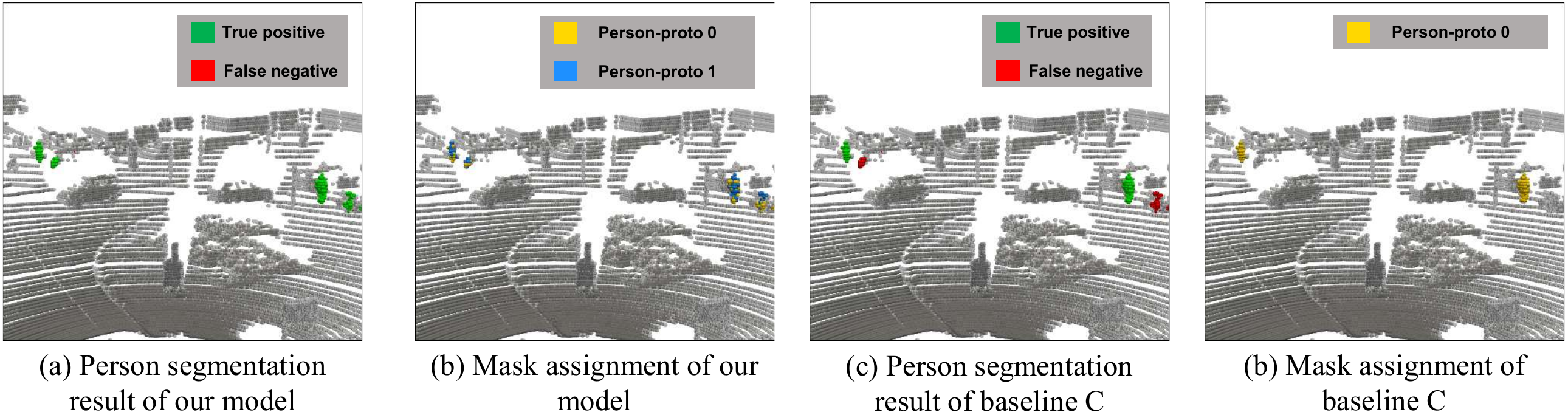}
    \caption{We visualize a segmentation result of the ``person" class. 
    (a) and (c) shows that our model is better than baseline C. (b) and (d) explains the observation: a single prototype cannot cover the high variance intra-class patterns.}
    \label{fig:example}
\end{figure}
\subsection{Results}

\begin{figure}[t]
\begin{minipage}[b]{.48\linewidth}
    \centering
    \begin{tabular}{c|cccc}
        \hline
                    & A & B & C & Full \\
        \hline
         PFEM & \checkmark & \checkmark & \checkmark & \checkmark \\ 
         T &  &\checkmark & \checkmark& \checkmark\\
         PBW & & &\checkmark & \checkmark\\
         PD & & & &\checkmark \\
         \hline
         mIoU &62.30 & 48.86 & 63.67 & \textbf{64.62}\\
         \hline
         
    \end{tabular}
    \captionof{table}{Model component ablation study on SemanticKITTI val-set. 
    }
    \label{tab:ablation}
\end{minipage}
\begin{minipage}[b]{.48\linewidth}
    \centering
    \includegraphics[width=5.0cm]{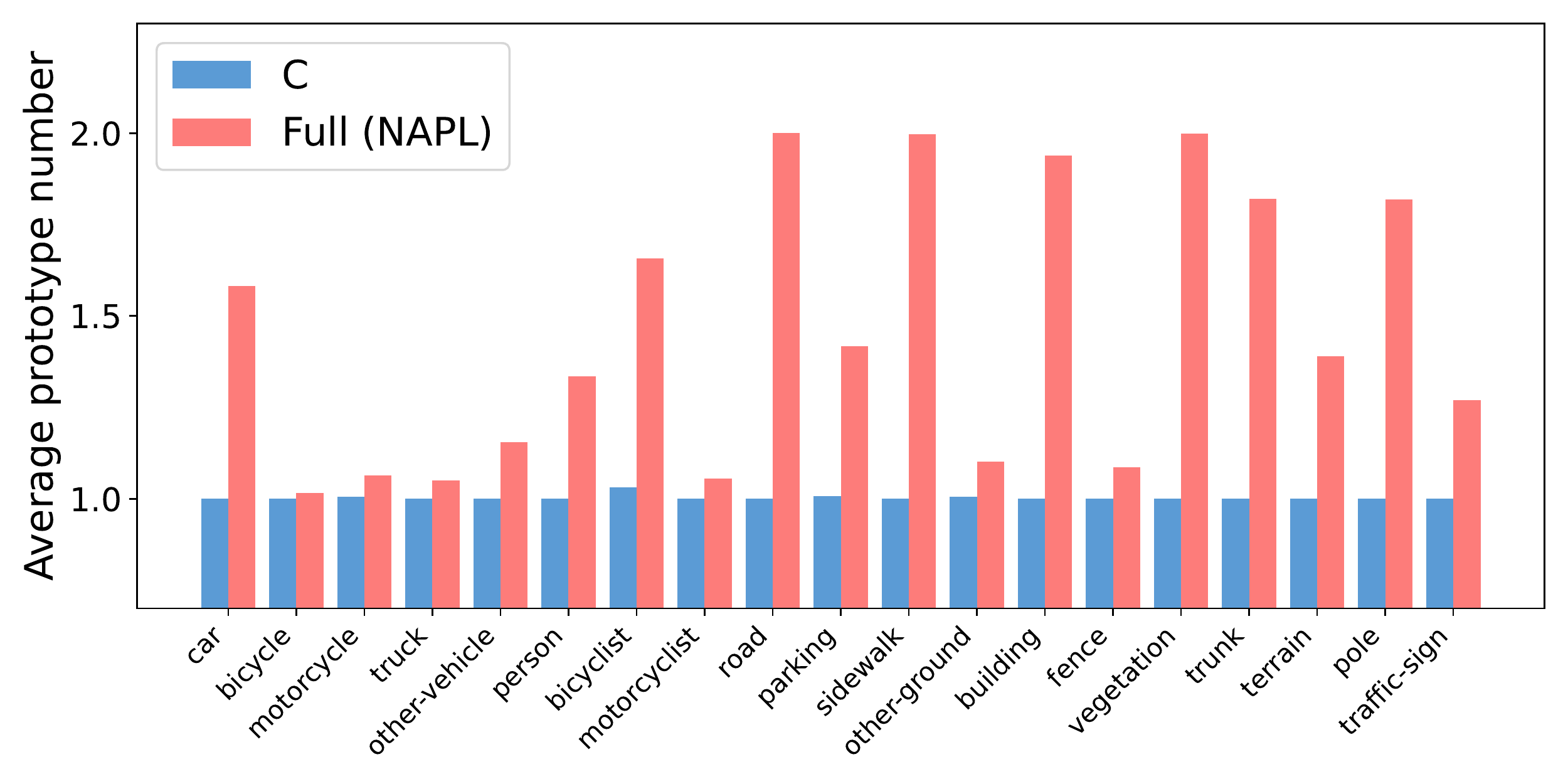}
    \captionof{figure}{Statistics of the average number of prototypes per class. 
    }
    \label{fig:class_static}
\end{minipage}

\end{figure}
We use the mean intersection of union (mIoU)~\cite{behley2019semantickitti} as our evaluation metric. The results are reported from both the validation and the test set of SemanticKITTI. 

\textbf{Quantitative evaluation.} In Table~\ref{tab:res-kiti}, we compare our number-adaptive prototype learning model with the existing 3D point cloud semantic segmentation models~\cite{qi2017pointnet,hu2020randla,thomas2019kpconv,xu2020squeezesegv3,milioto2019rangenet++,cortinhal2020salsanext} and the baseline point-wise classification model class-by-class. The result shows that our proposed number-adaptive prototype learning paradigm is better than the traditional point-wise classification paradigm. Specifically, in most of the classes where instances have different patterns including person, other-vehicle et al., our model has made significant improvements. 

\textbf{Ablation study.} In Table~\ref{tab:ablation}, we further study the effectiveness of individual components in our model, including the transformer decoder (T), pre-trained backbone weights (PBW), and the prototype dropout training strategy (PD). The results show that: i) directly adding a transformer module to the point cloud segmentation model and training them together greatly harms the performance by 13.44\%; ii) using a pre-trained backbone makes the model easy to train and boost the result by 1.37\%; and iii) the prototype dropout strategy can further promote the model performance
by 0.95\%. 

\textbf{Handling challenging cases.} In Fig.~\ref{fig:example}, we show a person segmentation case from the validation set to discuss the need for multiple prototypes for each class. Fig.~\ref{fig:example} (a) and (c) show the segmentation results of our model and baseline model C, respectively. The points in green are the true-positive points and the points in red are false-negative. Our model correctly segments all the points in this scene, while C misses the points belonging to the shorter person. Fig.~\ref{fig:example} (b) and (d) reveal the deeper 
reason: a single prototype cannot cover all the points of different persons, while two prototypes can describe the different patterns of people, and thus make a better segmentation. The visualization shows the superiority of our proposed number-adaptive prototype learning paradigm. 

\textbf{Prototype number analysis.}  Fig.~\ref{fig:class_static} presents the average number of prototypes for each class in each frame. With the prototype dropout strategy, our model can adaptively produce prototypes for each class, while the baseline model C using the same model architecture can only produce one prototype for each class. This result shows the effectiveness of our proposed training strategy.

\section{Conclusions}
In this paper, we propose a novel number-adaptive prototype learning paradigm for 3D point cloud semantic segmentation. To realize this, we leverage a transformer decoder in our model to learn prototypes for semantic categories. To enable training, we design a prototype dropout strategy to promote our model to produce number-adaptive prototypes for each class. The experimental results and visualization on SemanticKITTI demonstrate the effectiveness of our design.

\noindent\textbf{Acknowledgements.} This work is  supported by National Natural Science Foundation of China
under Grant 62171276, the Science and Technology Commission of Shanghai Municipal under Grant 21511100900 and CALT Grant 2021-01.
\clearpage
%
%
\bibliographystyle{splncs04}
\bibliography{ref}
\end{document}